\documentclass[10pt,twocolumn,letterpaper]{article}

\usepackage{iccv}
\usepackage{times}
\usepackage{epsfig}
\usepackage{graphicx}
\usepackage{amsmath}
\usepackage{amssymb}
\usepackage{soul}
\usepackage{enumitem}
\usepackage{multirow}
\usepackage[space, compress, sort]{cite}
\usepackage[accsupp]{axessibility}

\usepackage{booktabs} 
\usepackage{pifont}
\usepackage{overpic}
\newcommand{\cmark}{\ding{51}}%
\newcommand{\xmark}{\ding{55}}%

\usepackage[pagebackref=true,breaklinks=true,letterpaper=true,colorlinks,bookmarks=false ]{hyperref}
\hypersetup{citecolor=[RGB]{119,185,0}}

\iccvfinalcopy 

\def\iccvPaperID{5306} 

\ificcvfinal\fi

\begin{document}

\title{Fully Attentional Networks with Self-emerging Token Labeling}

\author{Bingyin Zhao\textsuperscript{1,2}\thanks{Work done during an internship at NVIDIA.}~~~Zhiding Yu\textsuperscript{1}\thanks{Corresponding author.}~~~Shiyi Lan\textsuperscript{1}~~Yutao Cheng\textsuperscript{3}~~Anima Anandkumar\textsuperscript{1,4}\\
Yingjie Lao\textsuperscript{2}~~Jose M. Alvarez\textsuperscript{1}\\
\textsuperscript{1}NVIDIA \quad \textsuperscript{2}Clemson University \quad \textsuperscript{3}Fudan University \quad \textsuperscript{4}Caltech
\and
}


\maketitle
\ificcvfinal\thispagestyle{empty}\fi

\begin{abstract}

Recent studies indicate that Vision Transformers (ViTs) are robust against out-of-distribution scenarios. In particular, the Fully Attentional Network (FAN) - a family of ViT backbones, has achieved state-of-the-art robustness. In this paper, we revisit the FAN models and improve their pre-training with a self-emerging token labeling (STL) framework. Our method contains a two-stage training framework. Specifically, we first train a FAN token labeler (FAN-TL) to generate semantically meaningful patch token labels, followed by a FAN student model training stage that uses both the token labels and the original class label. With the proposed STL framework, our best model based on FAN-L-Hybrid (77.3M parameters) achieves 84.8\% Top-1 accuracy and 42.1\% mCE on ImageNet-1K and ImageNet-C, and sets a new state-of-the-art for ImageNet-A (46.1\%) and ImageNet-R (56.6\%) without using extra data, outperforming the original FAN counterpart by significant margins. The proposed framework also demonstrates significantly enhanced performance on downstream tasks such as semantic segmentation, with up to 1.7\% improvement in robustness over the counterpart model. Code is available at \url{https://github.com/NVlabs/STL}.
\end{abstract}

\section{Introduction}
\label{sec:intro}
\begin{figure}[t]
    \centering
    \resizebox{0.48\textwidth}{!}{
    \includegraphics{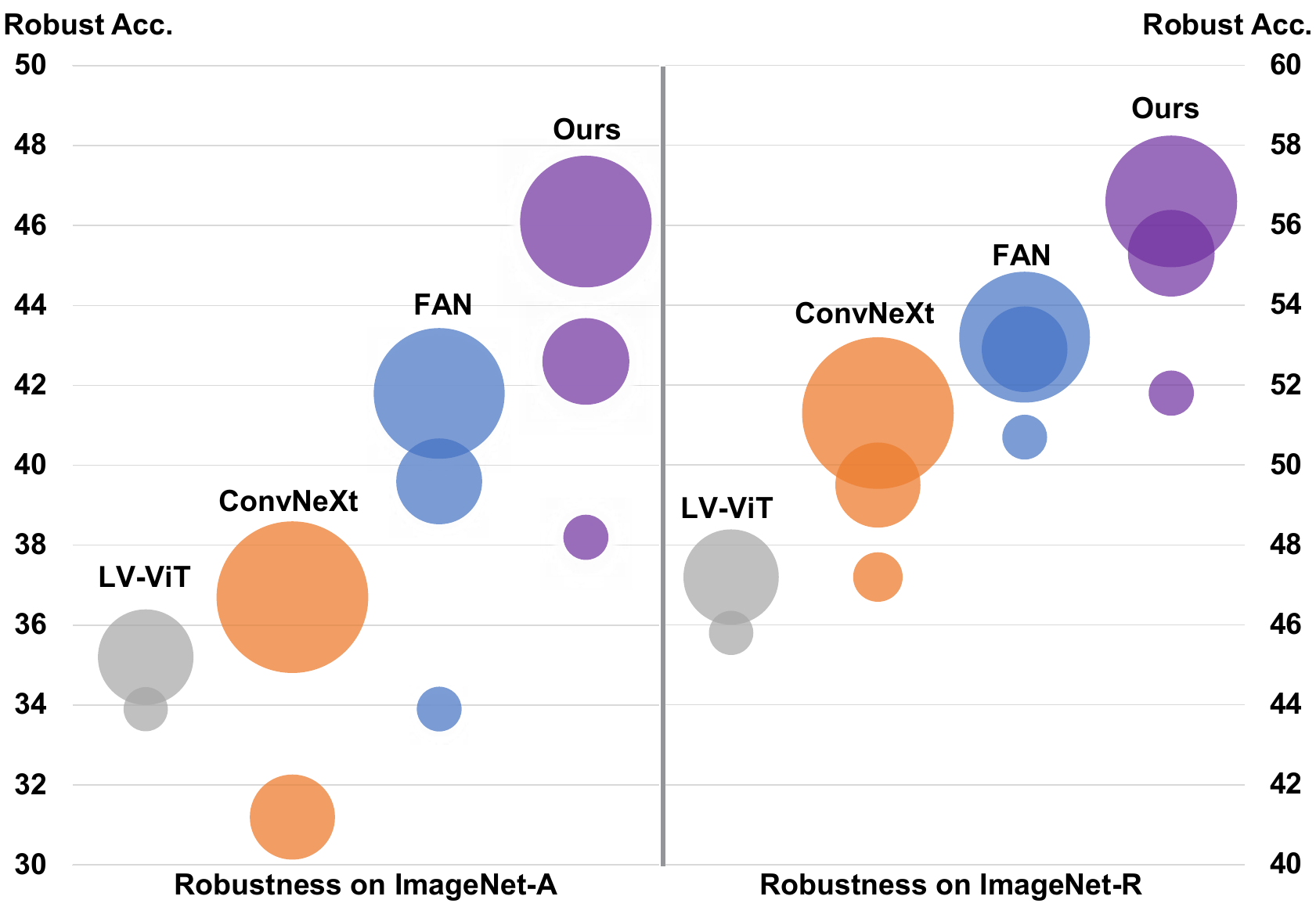}}
    \caption{\textbf{Results of zero-shot robustness against ImageNet-A and ImageNet-R.} Models trained on ImageNet-1K with self-emerging token labels from FAN show superior robustness to the out-of-distribution data. Our best model (with only 77.3M parameters) achieves robust accuracy of 46.1\% and 56.6\%  and sets a new record on ImageNet-A and ImageNet-R.}
    \label{fig:teaser}
\end{figure}

Vision Transformers (ViTs)~\cite{dosovitskiy2020image} have recently achieved remarkable success in visual recognition tasks. Such success is not only attributed to their self-attention representation but also to the newly developed training recipes. For instance, refinements in training techniques such as strong data augmentation and knowledge distillation~\cite{touvron2021training} greatly alleviate ViT's issue of being data-hungry and make them more accessible for training on ImageNet-1K. 


Another important development in the training recipe is token labeling~\cite{jiang2021all}, where patch tokens are assigned with labels to ViTs in a dense manner. In some sense, token labeling can also be considered as an alternative form of hard knowledge distillation. However, the dense nature of token labeling allows ViTs to leverage more fine-grained information in an image and take different categories and object localization into account. Compared to traditional knowledge distillation methods, token labeling enables ViTs to exploit a wider range of information in the image, leading to more accurate results. The success of token labeling depends on carefully-designed token-level annotators (i.e., token-labelers) that can provide accurate location-specific information (i.e., token labels) to patch tokens. In~\cite{jiang2021all}, this is done by a special re-labeling process~\cite{yun2021relabel} using convolutional neural networks (CNNs)~\cite{brock2021high} pre-trained on ImageNet-1K. While Vision Transformers have shown great promise in representation learning, less exploration has been conducted on modeling them as token-labelers. This raises two interesting questions: 
\begin{enumerate}
\vspace{-0.1em}
    \item Can Transformer-based models self-produce meaningful token labels?
    \item Can one improve the pre-training of ViTs with self-produced knowledge instead of external teachers? 
\end{enumerate}

\textbf{Our approach:} In this paper, we aim to answer the above questions. We propose a self-emerging token labeling (STL) framework that employs the self-produced token labels by ViT token-labelers instead of relying on CNNs. Our work is built on the recently proposed Fully Attentional Network (FAN)~\cite{luo2017understanding} for two reasons. 
First, FAN exhibits excellent self-emerging visual grouping on token features, which can be leveraged to generate high-quality token labels. Second, FAN is a family of ViT backbones with state-of-the-art accuracy and robustness. We aim to further improve this family of powerful backbones through a principled token-labeling design and validate its effectiveness. 
Our contributions can be summarized as follows:
\begin{itemize}[leftmargin=1.3em]
    \item Our work demonstrates that ViT models can be effective token-labelers. We propose a simple yet effective way to train a FAN token-labeler that can produce semantically meaningful token labels. 
    \item We perform an in-depth analysis and show critical contributors to the accuracy of token labels.  On top of the observations, we design a solution that retains more accurate token labels of the target object for improved model pre-training.
    \item Our models trained with STL set a new record on out-of-distribution datasets without using extra data than ImageNet-1K. Our best model achieves robust accuracy of 46.1\% on ImageNet-A and 56.6\% on ImageNet-R with only 77M parameters, as shown in Fig.~\ref{fig:teaser}.
    
    \item Experiments on downstream tasks demonstrate that the improved performance in backbone models is transferable to semantic segmentation and object detection.
\end{itemize}

Our STL framework is akin to the teacher-student training strategy introduced in knowledge distillation and consists of two stages:

\noindent\textbf{First stage:} We train a FAN token-labeler (FAN-TL) model to generate token-level annotations. Our task is essentially a ``chicken or the egg'' problem since there is no explicit supervision on how the token labels are generated. We tackle this by assigning supervising both the class token and the global average-pooled token. This produces semantically meaningful token labels as shown in Fig.~\ref{fig:token_confidence}(b).

\noindent\textbf{Second stage:} We train a FAN student model using the original class labels and the token labels from FAN-TL. 
Observing the imperfect quality of token labels, we introduce a token selection approach based on Gumbel-Softmax that adaptively selects tokens with high confidence. Labels of the selected tokens are of better quality and object grounding in general, leading to improved pre-training.
\section{Related Work}
\label{sec:background}

\subsection{Vision Transformers}
Vision Transformers~\cite{dosovitskiy2020image} are a family of visual recognition models built upon Transformers~\cite{vaswani2017attention}. ViT splits an input image into a series of small patches, projects each as an embedding (a.k.a patch token) and appends with position embeddings. The resulting patch tokens and an extra learnable class token that aggregates global information for classification are then fed into a sequence of Transformer encoders consisting of multi-head self-attention and FFN blocks. A linear projection layer is appended to the class token to predict the class probabilities. 


\subsection{Fully Attentional Networks}


Several concurrent works point out that ViTs exhibit excellent zero-shot robustness against out-of-distribution samples~\cite{naseer2021intriguing,bai2021transformers,paul2022vision,xie2021segformer}. Some works propose to use negative data augmentation~\cite{qin2021understanding} and adversarial training~\cite{herrmann2022pyramid,mao2022enhance} to further enhance the robustness. Recently, FAN~\cite{zhou2022understanding} was introduced as a family of ViT backbones with state-of-the-art accuracy and robustness. 
FAN inherits the self-attention blocks of plain ViT but additionally introduces a channel attention block that adopts an attention-based design that aggregates the cross-channel information in a more holistic manner, leading to improved representation.


\subsection{Token Labeling}
Token labeling~\cite{jiang2021all} has been proposed to improve ViT pre-training. From the perspective of training strategy, it is similar to knowledge distillation~\cite{yuan2020revisiting,touvron2021training} since both adopt a teacher-student mode. It is also related to ReLabel~\cite{yun2021re}, which provides images with multi-label annotations instead of single ones. However, both ReLabel and knowledge distillation depend on image-level labels as global supervision while token labeling assigns labels to each image patch token and supervises the student model in a dense manner. {Token labeling is also inherently related to tokenization in BEiT~\cite{bao2021beit}, where an offline pre-trained discrete VAE is employed as the tokenizer to encode patches into visual tokens (i.e., code from a visual codebook). Different from token labels, these visual tokens do not possess explicit semantically meanings since they originate from an unsupervisedly trained codebook.} Our method differs from prior token labeling and distillation methods where pre-trained convolutional neural networks are widely used as the token-labeler. Instead, our approach to unifies both teacher and student under homogeneous Vision Transformer architectures to generate high-quality token labels.


\subsection{Emerging Properties of ViTs}
It was found that the localization of objects emerges in image classification with CNNs. This interesting phenomenon, also known as class activation maps (CAM)~\cite{zhou2016learning}, lays the foundation for token labeling. Recent studies reveal that ViTs demonstrate excellent capability for object localization without explicit supervision. For instance, DINO~\cite{caron2021emerging} shows that self-supervised ViT features generate semantically meaning object segmentation. Methods like GroupViT~\cite{xu2022groupvit} show that semantic segmentation emerges in ViTs using only text supervision. Similarly, FAN reveals that the robustness of ViT models is correlated to their excellent visual grouping capability. This feature motivates us to develop self-emerging token labeling on top of the FAN models.

\section{Method}
\label{sec:method}


As mentioned, we propose a self-emerging token labeling (STL) framework that uses self-produced token labels to improve ViT pre-training. STL consists of two stages: 1) training an effective token-labeler, and 2) training a student model with self-emerging token labels.

In the first stage, we train a FAN token-labeler (FAN-TL) to generate high-quality token labels. As discussed in Sec.~\ref{sec:intro} and Sec.~\ref{sec:background}, FAN demonstrates strong robustness and capability to obtain semantically meaningful visual grouping that can correctly captures the object gestalt. These great features of FAN allow us to obtain high quality token labels without bells and whistles. In the second stage, we then train a FAN student model with image-level labels and self-emerging token labels from FAN-TL. At high level, FAN-TL and FAN student models follow the same architectural design as FAN but slightly modify the structure of patch tokens. We attach a linear layer to each patch token to accommodate token labels, similar to the classification head added to the class token in the original FAN and ViT design. The rest of this section describes the implementation details of the above two-stage pipeline.

\subsection{Training FAN Token-Labelers}

The original FAN employs the training paradigm that only takes image-level labels as supervision. We denote the input image as $\mathbf{I}$, the sequence of small patches as $[\mathbf{I}_{p_1},\mathbf{I}_{p_2}, \dots \mathbf{I}_{p_N}]$, the output of FAN encoder as $[\mathbf{T}_{cls}, \mathbf{T}_{p_1}, \dots, \mathbf{T}_{p_N}]$, where $N$ is the number of patch tokens, $\mathbf{T}_{cls}$ represents the class token and $\mathbf{T}_{p_1},\dots,\mathbf{T}_{p_N}$ represent the patch tokens, respectively. The training objective can be mathematically expressed as follows:
\begin{equation}
    \mathcal{L} = \mathcal{H}(\mathbf{T}_{cls}, \mathbf{Y}_{cls}),
    \label{equ:cls loss}
\end{equation}
where $\mathcal{H}(\cdot)$ is the softmax cross entropy loss and $\mathbf{Y}_{cls}$ is the image-level label of the corresponding class. 

\begin{figure}[htbp]
    \centering
    \resizebox{0.48\textwidth}{!}{
    \includegraphics{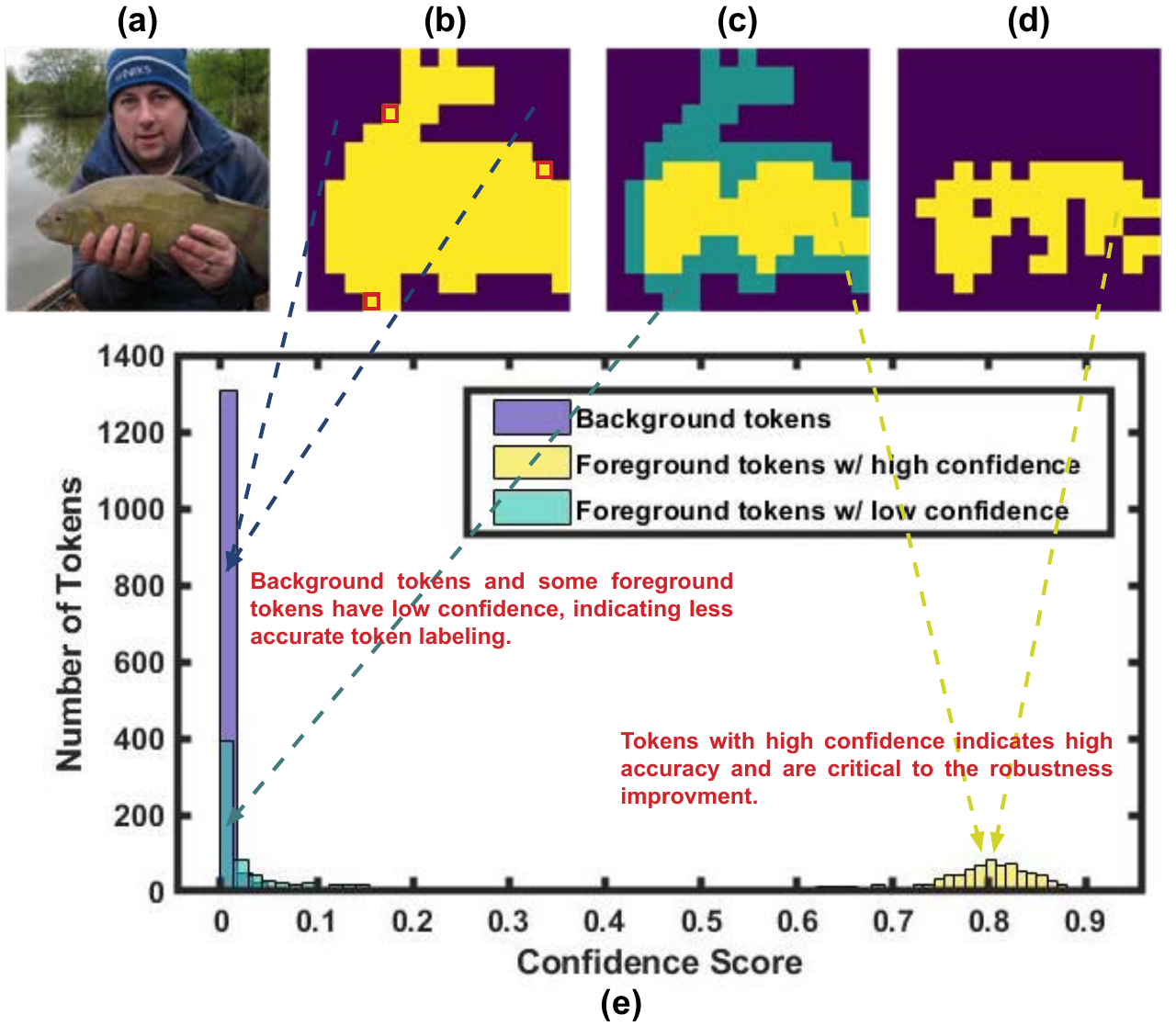}}
    \caption{\textbf{Illustration of token labels generated by FAN-TL and the token label confidence score distribution.} (a). original image (class: ``tench''), (b). binary color map of token labels (yellow: tokens classified as ``tench'', dark blue: tokens not classified as ``tench'') (c). trinary color map of token labels (cyan: \textit{foreground tokens} with low confidence, yellow: \textit{foreground tokens} with high confidence), (d). binary color map of \textit{foreground tokens} selected by Gumbel-Softmax, (e). token label confidence score distribution of a batch of 16 images.}
    \label{fig:token_confidence}
\end{figure}

The key to token labeling is the generation of accurate token labels that provide location-specific information. However, following the conventional training paradigm in Eq.~\ref{equ:cls loss}, token outputs of FAN models are not semantically well-guided since they are not supervised during training. We propose a simple yet effective method to address this issue. Our idea is inspired by the intriguing phenomenon in ViTs training that meaningful object segmentation naturally emerges~\cite{caron2021emerging}. Unlike the self-supervised training in DINO, we leverage FAN's strong capability of visual grouping~\cite{buhmann1999image} and devise a fully supervised approach that allows FAN to generate accurate and semantically meaningful token labels. We perform global average-pooling on all patch tokens and then simultaneously assign the class label to the class and the average-pooled tokens. The training objective of FAN-TL can be written as follows:
\vspace{-0.5em}
\begin{equation}
    \mathcal{L} = \mathcal{H}(\mathbf{T}_{cls}, \mathbf{Y}_{cls}) + \alpha\cdot\mathcal{H}(\frac{1}{N}\sum\limits_{i=1}^N\mathbf{T}_{p_i}, \mathbf{Y}_{cls}),
    \label{equ:FAN-TL loss}
\end{equation}
where $\alpha$ weights the importance of two loss functions. We set $\alpha$ to 1 in our experiments. We demonstrate a visualization example of the token labels generated by FAN-TL in Fig.~\ref{fig:token_confidence}(b). The yellow area (i.e., foreground) represents the tokens with the same labels as the image-level label (we term \textit{foreground tokens}). In contrast, the dark blue area (i.e., background) represents the tokens with different labels from the image-level label (we term \textit{background tokens}). It can be seen that the self-emerging token labels show a meaningful segmentation of the target object ``tench''.

\begin{figure*}[htbp]
    \centering
    \resizebox{1\textwidth}{!}{
    \includegraphics{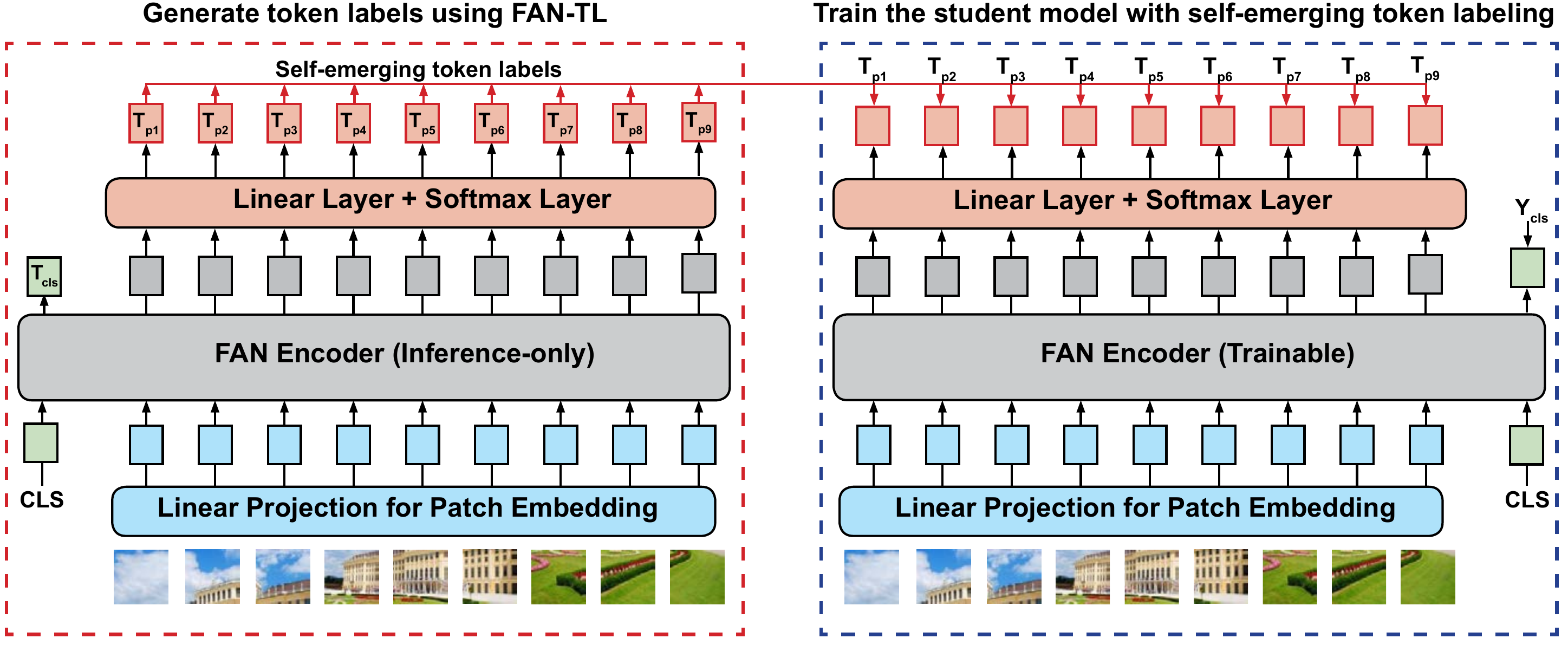}}
    \caption{\textbf{Illustration of Stage 2: Training student models with self-emerging token labels.} In the training, token labels are generated by FAN-TL and assigned to patch tokens of student models. We incorporate the token labels and class labels to train student models jointly. FAN-TL can self-identify the incorrect token labels upon the confidence score. Tokens with high confidence scores offer a more accurate segmentation of objects and are crucial to robustness improvement. By applying spatial-only data augmentation to the inputs and Gumbel-Softmax to the token outputs of FAN-TL, we obtain the most accurate and critical token labels.}
    \label{fig:FAN-V2}
\end{figure*}

\subsection{Training Student Models with STL}
\label{sec:stage2}
Training student models is straightforward. As illustrated in Fig.~\ref{fig:FAN-V2}, we take the token outputs of FAN-TL and assign them as the labels for patch tokens of student models. We then incorporate token labels with class labels and jointly optimize the loss on the class token and all patch tokens. The training objective is shown as follows:
\vspace{-0.5em}
\begin{equation}
    \mathcal{L} = \mathcal{H}(\mathbf{T}_{cls}, \mathbf{Y}_{cls}) + \beta\cdot\frac{1}{N}\sum\limits_{i=1}^N\mathcal{H}(\mathbf{T}_{p_i}, \mathcal{F}({\mathbf{I}}_{p_i}))
    \label{equ:FAN_v2_loss}
\end{equation}
where $\mathcal{F}(\cdot)$ represents the patch token outputs of FAN-TL and $\beta$ is the hyper-parameter to balance two loss functions. $\beta$ is set to 1 in our experiments. Note the correctness of token labels (especially the \textit{foreground tokens}) is critical since wrong labels introduce worthless local information. Thus, we propose several implementation tricks to generate and retain highly accurate token labels based on the following two observations.

\textbf{Observation 1: Data augmentations remarkably affect the accuracy of token labels.} Recent studies~\cite{touvron2021training,yuan2021tokens} indicate data augmentations improve the classification accuracy and robustness of ViT models. However, most augmentations undermine the accuracy of token labels. For example, augmentations in RandAug~\cite{cubuk2020randaugment}, such as posterize and solarize, alter pixel values, CutOut~\cite{zhong2020random} randomly erases a part of images,  Mixup~\cite{zhang2017mixup} and CutMix~\cite{yun2019cutmix} mix content from two images, which all make it difficult for FAN-TL to assign correct labels to patch tokens. To this end, we disable these data augmentations and keep spatial-only ones (i.e., flip, rotate, shear and translation) for FAN-TL when generating token labels.  Note that we still apply full data augmentations for the student model. The concrete settings of data augmentations are summarized in Table~\ref{tab:aug_setting}. Moreover, the strong data augmentations can be regarded as noises akin to corruptions and perturbations in out-of-distribution datasets. The input data augmentation discrepancy between FAN-TL and the student model enables the self-emerging token labels to provide clean and correct information regardless of the noise, thus improving the model's robustness.

\begin{table}[t]
\centering
\setlength{\tabcolsep}{3pt}
\resizebox{\linewidth}{!}{
\begin{tabular}{l|ccccc}
Model (Stage)   & Spatial & RandAug & CutOut & MixUp & CutMix \\ \midrule
FAN-TL (Stage1) &   \cmark      &   \cmark      &    \cmark    & \cmark      &  \cmark      \\ \midrule
Student (Stage2) &   \cmark      &   \cmark      &    \cmark    & \cmark      &  \cmark  \\ \midrule
FAN-TL (Stage2) &  \cmark       &  \xmark       &   \xmark     &  \xmark     &  \xmark      \\ \bottomrule
\end{tabular}
}
\caption{\textbf{Detailed settings of data augmentations.} We apply spatial-only data augmentations to inputs of FAN-TL in stage 2 to improve the accuracy of token labels.}
\label{tab:aug_setting}
\end{table}

\textbf{Observation 2: Not all the self-emerging token labels are correct, even with spatial-only data augmentations.} For example, as shown in Fig.~\ref{fig:token_confidence}(b), some patches (red squares) that only contain ``human'' and ``lake'' are misclassified as the target object ``tench'' by FAN-TL. It is particularly crucial to ensure the accuracy of foreground token labels as they contain the target object of the image that dominates the prediction result. However, it is challenging to determine which labels are incorrect due to the lack of patch-level ground truth. Interestingly, FAN-TL can self-identify these misclassified \textit{foreground tokens} according to the token label confidence score, which may also be attributed to its visual grouping ability. The token label confidence score is defined as the maximal class probability of each token output. We find that tokens with correct labels tend to have higher confidence scores than those with incorrect labels. As shown in Fig.~\ref{fig:token_confidence}(c) and Fig.~\ref{fig:token_confidence}(e), the yellow area indicates the \textit{foreground tokens} with high confidence scores ($0.7 \sim 0.9$). These tokens yield a highly accurate segmentation of the target object. Contrastly, the cyan area represents the \textit{foreground tokens} with low confidence scores ($0\sim 0.2$), which are coincidentally the ones with incorrect labels. 

We attempt to assign highly accurate labels to the target object (i.e., \textit{foreground tokens}). However, it is intractable to exhaustively examine the confidence score of each patch token for all the images as it significantly increases the training time and requires much more computational resources. Therefore, we propose a lightweight alternative by applying the Gumbel-Softmax~\cite{DBLP:conf/iclr/JangGP17} on the top of token outputs. Mathematically, it can be expressed as:
\begin{equation}
    \mathbf{y}_i = \frac{e^{(log(\pi_i)+\mathcal{G}_i)/\tau)}}{\sum_{j=1}^{k}e^{(log(\pi_j)+\mathcal{G}_j)/\tau)}}
\end{equation}
where $\mathbf{y}$ is the $k$-dimension softmax vector, $\pi$ are class probabilities, $\mathcal{G} \sim$ $Gumbel(0,1)$ are i.i.d. samples drawn from the standard Gumbel distribution and $\tau$ is the softmax temperature. Token labels with high confidence scores remain unchanged after applying Gumbel-Softmax, while labels with low confidence scores are highly likely to change. As shown in Fig.~\ref{fig:token_confidence}(d), we preserve the correct token labels and eliminate the incorrect ones in a simple yet effective way, achieving high accuracy of \textit{foreground tokens} labels and more precise segmentation of the object. Furthermore, since the training objective of the patch tokens side can be considered a self-training process~\cite{mclachlan1975iterative,rosenberg2005semi,xie2020self}, we convert the softmax outputs to ``one-hot'' probability distribution (i.e., hard labels) as ~\cite{DBLP:conf/nips/SohnBCZZRCKL20} shows that the use of hard labels in self-training encourages the model's predictions to be high-confidence via entropy minimization~\cite{grandvalet2004semi}. 
Following the aforementioned spatial-only data augmentation, we can then rewrite the training objective of the student model:
\vspace{-0.5em}
\begin{equation}
    \mathcal{L} = \mathcal{H}(\mathbf{T}_{cls}, \mathbf{Y}_{cls}) + \beta\cdot\frac{1}{N}\sum\limits_{i=1}^N\mathcal{H}(\mathbf{T}_{p_i}, \hat{\mathcal{F}}(\hat{\mathbf{I}}_{p_i}))
    \label{equ:FAN_v2_loss_final}
\end{equation}
where $\hat{\mathcal{F}}(\cdot)$ is the one-hot encoded Gumbel-Softmax outputs of FAN-TL and $[\hat{\mathbf{I}}_{p_1}\dots\hat{\mathbf{I}}_{p_N}]$ are image patches with the spatial-only data augmentation. Meanwhile, we notice that all \textit{background tokens} (i.e., the dark blue area) also have low confidence scores. This is probably because these tokens neither contain apparent features associated with the target object nor clear features related to any other class in the dataset. Nevertheless, we keep all tokens (even those with low confidence scores) in training, following the practice in~\cite{jiang2021all} as involving more tokens for loss computation yields better performance.


\section{Experiments}
\label{sec:exp}

\subsection{Datasets and Evaluation Metrics}

We evaluate our method on the image classification task and its transferability to downstream semantic segmentation and object detection tasks.


\textbf{Datasets.} For image classification, we test model performance and robustness on ImageNet-1K (IN-1K)~\cite{deng2009imagenet}, ImageNet-C (IN-C)~\cite{hendrycks2019benchmarking}, ImageNet-A (IN-A)~\cite{hendrycks2021natural} and ImageNet-R (IN-R)~\cite{DBLP:conf/iccv/HendrycksBMKWDD21}. IN-C contains natural corruptions from noise, blur, weather, and digital categories and is widely used to evaluate model's robustness against shifted distribution data. IN-A and IN-R consist of images with different distributions from ImageNet training distribution, such as natural adversarial examples and pictures generated by artistic rendition, thus are widely used to measure the robustness against out-of-distribution data. For semantic segmentation and object detection, we evaluate models on Cityscapes (City)~\cite{Cordts2016Cityscapes}, Cityscapes-C (City-C) and COCO~\cite{cocodataset}. Similar to IN-C, City-C has corruptions from the same four categories.

\textbf{Metrics.} We adopt standard evaluation metrics for image classification: clean accuracy for IN-1K and robust accuracy for IN-C, IN-A and IN-R. We also report mean corruption error (mCE)~\cite{hendrycks2019benchmarking} on IN-C. For semantic segmentation and object detection, we  evaluate the model performance using the clean and robust mean Intersection over Union (mIoU) on City and City-C and the mean average precision (mAP) on COCO. Additionally, we use retention rate as the metric to reflect the resilience of the model robustness and fairly compare models with different capacities. The retention rate is defined as $R= \frac{\text{Robust Acc.}}{\text{Clean Acc.}}$.

\subsection{Implementation Details}

Experiments are conducted on 8 NVIDIA Tesla V100s and codes are built upon Pytorch~\cite{paszke2019pytorch}, timm~\cite{rw2019timm} library and MMSegmentation~\cite{contributors2020mmsegmentation} toolbox. We adopt FAN-Hybrid as the model architecture for FAN-TL and student models. For image classification, we train the models on ImageNet-1K using AdamW optimizer with a learning rate of 4e-3 and batch size of 2048 for 350 epochs. We employ the cosine scheduler with a decay rate of 0.1 to adjust the learning rate every 30 epochs. The loss weight $\alpha$ in Eq.~\ref{equ:FAN-TL loss} and $\beta$ in Eq.~\ref{equ:FAN_v2_loss_final} are set to 1. We apply spatial, RandAug, CutOut, Mixup and CutMix data augmentation in the training and a label smoothing ratio of 0.9 to class and token labels. As discussed in Sec.~\ref{sec:method}, we apply spatial-only data augmentations on inputs of FAN-TL and Gumbel-Softmax on patch token outputs to obtain accurate token labels. We employ pre-trained image classification models for semantic segmentation as encoders and the SegFormer~\cite{xie2021segformer} head as the decoder. We follow the same training recipe as SegFormer and train our models on Cityscapes using AdamW with a learning rate of 6e-5 and a batch size of 8 for 160K iterations. The learning rate scheduler is set to ``poly'' with a default factor of 1.0. Random resizing, flipping, and cropping are applied as data augmentations in training. For object detection, we follow the same practice as Swin Transformer~\cite{liu2021swin} + Cascade Mask R-CNN~\cite{cai2018cascade} and employ AdamW (initial learning rate of 1e-4, weight decay of 0.05, and batch size of 16) to train our models on COCO for 36 epochs.

\begin{table}[tbp]
    \small
    \centering
    \setlength{\tabcolsep}{3pt}
    \resizebox{\linewidth}{!}{
    \begin{tabular}{l|cccc}  
    Model  &  Param./FLOPs   & IN-1K       & IN-C    &  Retention  
    \\
    \midrule
    ResNet18 \cite{he2016deep} & 11M/1.8G & 69.9  & 32.7 & 46.8\%     
    \\  
    MBV2 \cite{sandler2018mobilenetv2}  &  4M/0.4G  & 73.0 &  35.0  & 47.9\% 
     \\ 
    EffiNet-B0 \cite{tan2019efficientnet}  &  5M/0.4G  & 77.5 &  41.1  & 53.0\%  
     \\ 
    PVTV2-B0 \cite{wang2021pyramid}  &  3M/0.6G  & 70.5 &  36.2  & 51.3\% 
     \\ 
    PVTV2-B1 \cite{wang2021pyramid}  &  13M/2.1G  & 78.7 &  51.7  &  65.7\% 
     \\ 
     LV-ViT-T \cite{jiang2021all}  &  9M/2.1G  & 79.1 & 51.6   &  65.2\% 
     \\

         FAN-T-Hybrid \cite{zhou2022understanding} &  7M/3.5G  & 80.1 &   57.4 & 71.4\%
     \\ 
         STL (FAN-T-Hybrid)  &  8M/3.6G  & 79.9 & \textbf{58.2}   &  \textbf{72.8\%}
     \\ 
     \midrule
    ResNet50 \cite{he2016deep}  &  25M/4.1G  & 79.0 &  50.6  & 64.1\% 
     \\ 
         DeiT-S \cite{touvron2021training}  &  22M/4.6G  & 79.9 &  58.1  & 72.7\% 
     \\ 
         Swin-T \cite{liu2021swin}  &  28M/4.5G  & 81.3 &  55.4  & 68.1\% 
     \\ 
         ConvNeXt-T \cite{liu2022convnet}  &  29M/4.5G  & 82.1 &  59.1  & 71.9\% 
     \\ 
     LV-ViT-S \cite{jiang2021all}  &  26M/6.6G  & 83.3 & 59.7   & 71.7 \% 
     \\ 
     
         FAN-S-Hybrid \cite{zhou2022understanding} &  26M/6.7G  & 83.5 &  64.7  & 77.5\% 
     \\ 
             STL (FAN-S-Hybrid)  &  27M/6.8G  & 83.4 &  \textbf{65.5}  & \textbf{78.5\%} 
     \\ 
    \midrule
         Swin-S \cite{liu2021swin}  &  50M/8.7G  & 83.0 &  60.4  & 72.8\% 
     \\ 
         ConvNeXt-S \cite{liu2022convnet}  &  50M/8.7G  & 83.1 & 61.7   & 74.2\% 
     \\ 
     LV-ViT-M \cite{jiang2021all}  &  56M/16.0G  & 84.0 &  62.0  &  73.8\% 
     \\ 
     
         FAN-B-Hybrid \cite{zhou2022understanding} &  50M/11.3G  & 83.9 &  66.4  & 79.1\% 
     \\ 
         STL (FAN-B-Hybrid)  &  51M/11.4G  & 84.5 &  \textbf{68.2}  & \textbf{80.7\%} 
     \\ 
     
    \midrule
         DeiT-B \cite{touvron2021training}  &  89M/17.6G  & 81.8 &  62.7  &  76.7\% 
     \\ 
         Swin-B \cite{liu2021swin}  &  88M/15.4G  & 83.5 &  60.4  & 72.3\% 
     \\ 
         ConvNeXt-B \cite{liu2022convnet}  &  89M/15.4G & 83.8  & 63.0 &  75.2\%
     \\ 
     
         FAN-L-Hybrid \cite{zhou2022understanding}  &  77M/16.9G  & 84.3 &  68.3  & {81.0\%} 
     \\ 
         STL (FAN-L-Hybrid) &  77M/17.0G  & 84.7 &  \textbf{68.8}  & \textbf{81.2\%} 
     \\ 
    \bottomrule
    \end{tabular}
    }
    \caption{\textbf{Results on image classification.} We report clean and robust accuracy on ImageNet-1K and ImageNet-C. Retention rate is defined as $\frac{\text{Robust Acc.}}{\text{Clean Acc.}}$. LV-ViTs are vanilla ViTs trained with a CNN token-labeler. Our models trained with STL achieve superior robustness and retention rate in all cases. Meanwhile, our method also improves the clean accuracy of models with a larger capacity (e.g., FAN-B-Hybrid and FAN-L-Hybrid).}
    \label{tab:robustness_imagenet_c}

\end{table}

\subsection{Results on Image Classification} 
\label{sec:image_classification_results}
We first show the performance of models trained with STL on the image classification task and compare them with other SOTA models in Table~\ref{tab:robustness_imagenet_c}. To evaluate the zero-shot robustness against the distributional shift, all models are trained on ImageNet-1K data and directly used for inference on ImageNet-C without finetuning. We use the same model type for FAN-TL and the student model. It can be seen that Transformer-based models are more robust than CNN-based models in general. At all size levels, our models show superior robust accuracy and retention rate to other models, including the original FAN models trained solely with the class labels, indicating the effectiveness of STL in improving model robustness. Notably, our models surpass LV-ViTs~\cite{jiang2021all} (i.e., vanilla ViT models trained with a CNN token-labeler) in both clean and robust accuracy by significant margins, which validates the importance of channel attention block in FAN and reveals the potential of self-emerging token labels from Transformer-based models.

\begin{table}[tbp]
\centering
\small
\setlength{\tabcolsep}{3pt} 
\resizebox{\linewidth}{!}{
\begin{tabular}{l|ccccc}
Model             & Params (M)           & Clean                & IN-A                 & IN-R                 & \multicolumn{1}{c}{mCE} ($\downarrow$)\\ \midrule
Swin-T~\cite{dong2021cswin}             &     28.3                 &      81.2                &     21.6                 &      41.3                &     59.6                     \\

ConvNext-T~\cite{liu2022convnet}       & 28.6  &  82.1          &  24.2     &  47.2    &  53.2          \\

RVT-S~\cite{mao2021towards}             &     23.3                 &     81.9                 &     25.7                 &      47.7                &     51.4                     \\

XCiT-S12~\cite{el2021xcit}         &    26.3                  &    81.9                  &       25.0               &       45.5               &       51.5                   \\

LV-ViT-S \cite{jiang2021all} &      26.0      &  83.3     &  33.9    &  45.8    &  52.9    \\

FAN-S-Hybrid \cite{zhou2022understanding} &      26.3      &  83.5     &  33.9    &  50.7    &  47.8    \\ 

STL (FAN-S-Hybrid) &    26.5        &  83.4   &  \textbf{38.2}    &  \textbf{51.8}    &  \textbf{47.3}   \\ \midrule

Swin-S~\cite{dong2021cswin}             &      50.0                &       83.4               &       35.8               &       46.6               &      52.7                    \\

ConvNext-S~\cite{liu2022convnet}        & 50.2 &   82.1         &  31.2     &  49.5    &  51.2      \\

XCiT-S24~\cite{el2021xcit}          &    47.7                  &      82.6                &     27.8                 &       45.5               &     49.4                     \\ 

LV-ViT-M \cite{jiang2021all} &      56.0     &  84.0     &  35.2    &  47.2    &    50.5  \\    

FAN-B-Hybrid \cite{zhou2022understanding} &      50.4      &   83.9    & 39.6     &   52.9   &  45.2    \\

STL (FAN-B-Hybrid) &    50.9       &   {84.5}    &   \textbf{42.6}   &  \textbf{55.3}    &  \textbf{43.6}    \\ \midrule

Swin-B~\cite{dong2021cswin}            &  87.8      &  {83.4}       & 35.8      & 46.6     & 54.4                          \\

MAE-ViT-B~\cite{he2022masked}     &      86.0               &   83.6                   &     35.9                 &   48.3                   &      51.7                    \\

ConvNext-B~\cite{liu2022convnet}        &    88.6        &   83.8    &  36.7    &  51.3    & 46.8     \\

RVT-B~\cite{mao2021towards}             &     91.8                 &     82.6                &      28.5                &      48.7                & 46.8                           \\ 

DAT-AugReg-ViT~\cite{mao2022enhance}    &     86.0                &     81.5                &      30.2                &      47.3                & 44.7                           \\ 

FAN-L-Hybrid \cite{zhou2022understanding} &      76.8      &  84.3     & 41.8     &  53.2    &  43.0    \\ 

STL (FAN-L-Hybrid) &    77.3        & {84.7}      &  \textbf{46.1}    &  \textbf{56.6}    & \textbf{42.5}     \\ \bottomrule
\end{tabular}}
\caption{\textbf{Results on out-of-distribution datasets.} We report the mean corruption error (mCE) for ImageNet-C, a lower value indicates better robustness. The improved robustness of models trained with STL is well generalized to out-of-distribution datasets and achieves even better robustness on ImageNet-A and ImageNet-R.}
\label{ood_generalization}
\end{table}

\subsection{Robustness against Out-of-distribution Data} 
\label{sec:ood_results}
Token labels embed rich local information of image patches. We adopt spatial-only data augmentation and Gumbel-Softmax on FAN-TL to retain highly accurate labels for foreground tokens to ensure that self-emerging token labels always provide correct information for student models. The practice promotes the generalization performance of student models as they can make robust predictions even with different input data distributions. To verify this, we then evaluate model robustness against out-of-distribution data, and results are summarized in Table~\ref{ood_generalization}. Similarly, models are not fine-tuned for testing. We find that LV-ViTs and original FAN models generalize well to out-of-distribution data, while other Transformer-based models and the SOTA CNN-based ConvNext models perform weaker. Despite the impressive performance of original FAN models, models trained with STL demonstrate an even better generalization ability and outperform all other models. The performance gains on IN-A and IN-R are more significant than IN-C and set a new state-of-the-art, indicating that the accurate self-emerging token labels are crucial to robustness against out-of-distribution data.

\begin{table}[tbp]
    \small
    \centering
    \setlength{\tabcolsep}{3pt}
    \resizebox{\linewidth}{!}{
    \begin{tabular}{l|cccc}
    Model  &  Encoder Size   & City       & City-C    &  Retention 
    \\
    \midrule
    DeepLabv3+ (R50)~\cite{kamann2020benchmarking}  &  25.4M  & 76.6 &  36.8  & 48.0\% 
     \\ 
    DeepLabv3+ (R101)~\cite{kamann2020benchmarking}  &  47.9M  & 77.1 &  39.4  &  51.1\%
     \\ 
    DeepLabv3+ (X65)~\cite{kamann2020benchmarking}  & 22.8M  & 78.4 &  42.7  &  54.5\%
     \\ 
    DeepLabv3+ (X71)~\cite{kamann2020benchmarking}  & -  & 78.6 &  42.5  &  54.1\%
     \\ 
     \midrule
    ICNet (\cite{zhao2018icnet})  &  -  & 65.9 &  28.0  & 42.5\% 
     \\ 
         FCN8s (\cite{long2015fully})  &  50.1M  & 66.7 &  27.4  & 41.1\% 
     \\ 
         DilatedNet (\cite{yu2015multi})  &  -  & 68.6 &  30.3  &  44.2\%
     \\ 
         ResNet38 (\cite{wu2019wider})  &  -  & 77.5 &  32.6  &  42.1\% 
     \\ 
         PSPNet (\cite{zhao2017pyramid})  &  13.7M  & 78.8 &  34.5  &  43.8\%
     \\ 
         ConvNeXt-T (\cite{liu2022convnet})  &  29.0M  & 79.0 &  54.4  &  68.9\% 
     \\ 
     \midrule
         SETR (\cite{heo2021rethinking}) &  22.1M  & 76.0 &  55.3  & 72.8\%  
     \\ 
         SWIN-T (\cite{liu2021swin})  &  28.4M  & 78.1 &  47.3  &  60.6\%
     \\ 
         SegFormer-B0 (\cite{xie2021segformer})  &  3.4M  & 76.2 &  48.8  &  64.0\%
     \\ 
         SegFormer-B1 (\cite{xie2021segformer})  & 13.1M  & 78.4 &  52.7  &  67.2\%
         \\
        SegFormer-B2 (\cite{xie2021segformer})  &  24.2M  & 81.0 &  59.6  &  73.6\%
     \\ 
         SegFormer-B5 (\cite{xie2021segformer})  &  81.4M  & 82.4 &  65.8  &  79.9\%
     \\ 
     \midrule
         FAN-B-Hybrid~\cite{zhou2022understanding}   &  50.4M  & 82.2 &  66.9  &  81.5\%
     \\ 
        STL (FAN-B-Hybrid)   &  50.9M  & \textbf{\textbf{\textbf{\textbf{82.5}}}} &  \textbf{\textbf{\textbf{68.6}}}  &  \textbf{83.2}\%
     \\ 
         \midrule
         FAN-L-Hybrid~\cite{zhou2022understanding}   & 76.8M &  82.3  & 68.7   & {83.5\%} 
     \\
         STL (FAN-L-Hybrid)   & 77.3M &  \textbf{82.8}  & \textbf{\textbf{69.2}}   & \textbf{83.6\%} 
     \\ 
    \bottomrule
    \end{tabular}}
    \caption{\textbf{Results on semantic segmentation.} We use mIoU as the evaluation metric. `R-' and `X-' refer to ResNet and Xception, respectively. Models trained with STL demonstrate an impressive transferability to the downstream task and achieve significantly better mIoU than other models on Cityscapes and Cityscapes-C.}
    \label{tab:robustness_cityscape_c}  
\vspace{-1em}
\end{table}
\vspace{-0.5em}

\subsection{Transferability to Semantic Segmentation} 
~\cite{he2019bag} shows pre-trained models using different training recipes perform differently in downstream tasks. ~\cite{jiang2021all} validates that token labeling with CNN token-labelers benefits semantic segmentation and improves the clean mIoU. We also evaluate the transferability of STL to semantic segmentation. As shown in Table~\ref{tab:robustness_cityscape_c}, pre-trained models with STL reveal better transferability than original FAN counterparts and other prestigious backbones. Remarkably, our approach achieves superior results on both clean and corrupted datasets.  As far as we know, this is the first work to reveal that applying dense supervision in backbone pre-training improves not only the clean performance but also the robustness of the downstream task.

\begin{table}[htbp]
\setlength{\tabcolsep}{10pt}
\resizebox{\linewidth}{!}{
\begin{tabular}{lcc}
\multicolumn{1}{l|}{Model}              & \multicolumn{1}{l}{Encoder Size} & \multicolumn{1}{l}{COCO (mAP)} \\ \midrule
\multicolumn{3}{c}{Cascade Mask-RCNN 3$\times$ schedule}                                                             \\ \midrule
\multicolumn{1}{l|}{ResNet-50~\cite{he2016deep}}          & 25M                              & 46.3                             \\
\multicolumn{1}{l|}{ResNeXt-101-32~\cite{xie2017aggregated}}     & -                                & 48.1                             \\
\multicolumn{1}{l|}{ResNeXt-101-64~\cite{xie2017aggregated}}    & -                                & 48.3                             \\
\multicolumn{1}{l|}{Swin-T~\cite{liu2021swin}}             & 28M                              & 50.4                             \\
\multicolumn{1}{l|}{ConvNeXt-T~\cite{liu2022convnet}}         & 29M                              & 50.4                             \\
\multicolumn{1}{l|}{FAN-S-Hybrid~\cite{zhou2022understanding}}       & 26M                              & 53.3                             \\
\multicolumn{1}{l|}{STL (FAN-S-Hybrid)} & 26M                              & \textbf{53.4}                             \\ \hline
\multicolumn{1}{l|}{Swin-S~\cite{liu2021swin}}             & 50M                              & 51.9                             \\
\multicolumn{1}{l|}{ConvNeXt-S~\cite{liu2022convnet}}         & 50M                              & 51.9                             \\
\multicolumn{1}{l|}{FAN-B-Hybrid~\cite{zhou2022understanding}}       & 50M                              & 53.5                             \\
\multicolumn{1}{l|}{STL (FAN-B-Hybrid)} & 50M                              & \textbf{53.9}                             \\ \midrule
\multicolumn{1}{l|}{Swin-B~\cite{liu2021swin}}             & 88M                              & 51.9                             \\
\multicolumn{1}{l|}{ConvNeXt-B~\cite{liu2022convnet}}         & 89M                              & 52.7                             \\
\multicolumn{1}{l|}{FAN-L-Hybrid~\cite{zhou2022understanding}}       & 77M                              & \textbf{54.1}                             \\
\multicolumn{1}{l|}{STL (FAN-L-Hybrid)} & 77M                              & \textbf{54.1}                             \\ \midrule
\end{tabular}}
\caption{\textbf{Results on object detection.} Models trained with STL outperform most CNN-based and transformer-based backbones. Even compared with original FAN models,  our models achieve at least on par or even better mAP on COCO, indicating our method can also benefit object detection.}
\label{tab:object_detection}
\vspace{-0.5em}
\end{table}

\subsection{Transferability to Object Detection}

We also validate STL's transferability to the object detection task on COCO and present the results in Table~\ref{tab:object_detection}. Models trained with STL outperform most CNN-based and Transformer-based backbones. Even compared with original FAN models, STL brings a notable performance gain for FAN-B while achieving comparable mean average precision on FAN-L and FAN-S. We notice the overall improvement in object detection is not as good as in image classification and semantic segmentation, possibly because the semantic segmentation task is more similar to token labeling from the perspective of dense prediction, while object detection involves different techniques such as regression~\cite{he2017mask,ren2015faster} and multi-stage refinement~\cite{he2017mask, cai2018cascade}.

\section{Ablation Study}
\label{sec:ablation}

\subsection{Impacts of Different Data Augmentation}
\begin{figure}[tbp]
\centering
\resizebox{0.3\textwidth}{!}{
\includegraphics{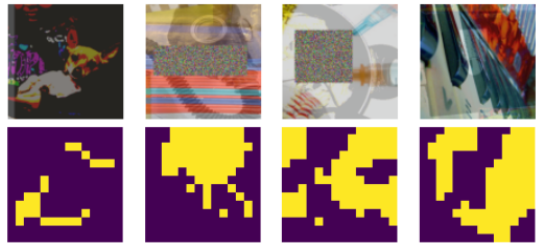}}
\caption{\textbf{Visualization results of token labels generated by FAN-TL with full data augmentations.} Strong augmentations significantly affect the quality of token labels.}
\label{fig:tl_with_strong_aug}
\end{figure}

We apply spatial-only data augmentation on FAN-TL to obtain accurate token labels that provide clean and correct information toward student models, which is vital to model robustness. {The strategy is a ``clean teacher noisy student'' design, where the student still uses all the augmentations. Our motivation behind this design is to let the teacher and student spatially aligned, with the teacher being as clean as possible to generate high quality token labels (Strong augmentations make it harder to generate good token labels as shown in Fig.~\ref{fig:tl_with_strong_aug}).} We compare the impacts of different data augmentation imposed on FAN-TL in Table~\ref{ablation_agumenation}. Experiments are conducted on the FAN-S-Hybrid model. The student model trained with token labels generated by spatial-only data augmentation achieves the best robustness while applying stronger augmentations harms the robust accuracy. Interestingly, applying the consistent full data augmentation on FAN-TL and the student model yields better clean accuracy, which reveals that different combinations of data augmentation may play different roles in improving model robustness and clean performance.

\begin{table}[tbp]
\centering
\small

\setlength{\tabcolsep}{3pt}
\resizebox{\linewidth}{!}{
\begin{tabular}{ccccccc}
\toprule
Spatial & RandAug & CutOut & MixUp & CutMix & IN-1K & IN-C \\ \midrule
\cmark        &  \xmark       &  \xmark      &  \xmark     &   \xmark     &  83.4     &   \textbf{65.5}   \\
\cmark        &   \cmark       &   \cmark      &  \xmark      &   \xmark     &  82.6     &  63.5    \\
\cmark         &   \cmark       &  \cmark       &   \cmark     &   \cmark      &  \textbf{83.6}     &   64.9   \\ \bottomrule
\end{tabular}
}
\caption{\textbf{Ablation on impact to the student model using various data augmentations on FAN-TL.} The different data augmentations are only imposed on the token-labeler's inputs, while the student model's inputs are fully augmented. Strong data augmentations on FAN-TL undermine token label accuracy and significantly affect the student model's robust accuracy.}
\label{ablation_agumenation}
\vspace{-0.5em}
\end{table}

\subsection{Impacts of Gumbel-Softmax}

We propose to use Gumbel-Softmax as a lightweight solution to retain correct token labels for foreground tokens and eliminate incorrect ones, which yields a more accurate segmentation of the object. We evaluate the impact of Gumbel-Softmax using the FAN-S-Hybrid model in Table~\ref{tab:gumbel-softmax}. As can be seen, models trained with token labels generated by Softmax and Gumbel-Softmax achieve comparable clean accuracy while Gumbel-Softmax improves robust accuracy. This further validates that accurate foreground token labels are critical to model robustness.


\begin{table}[htbp]
\centering
\small
\label{gumbel_softmax}
\setlength{\tabcolsep}{3pt}
\resizebox{\linewidth}{!}{
\begin{tabular}{l|ccccc}
Token Labeling & IN-1K & IN-C & mCE ($\downarrow$) & IN-A & IN-R \\ \midrule
Softmax & 83.4 & 65.2 & 47.6 & 37.2 & 51.7 \\ 
Gumbel-Softmax & 83.4 & \textbf{65.5} & \textbf{47.3} & \textbf{38.2} & \textbf{51.8} \\ \bottomrule
\end{tabular}
}
\caption{\textbf{Comparison of Softmax and Gumbel-Softmax.} Both methods achieve comparable clean accuracy while token labels generated by Gumbel-Softmax yield better robustness.}
\label{tab:gumbel-softmax}
\vspace{-0.5em}
\end{table}

\subsection{Training with Heterogeneous Token-Labelers}
In the previous experiments, we train student models with isomorphic FAN token-labelers (e.g., FAN-TL-S-Hybrid for FAN-S-Hybrid). We are interested in the impacts of training with a heterogeneous token-labeler and conduct the ablation in Table~\ref{tab:heterogeneous_token_labeler}. For each student model, we train with three different FAN token-labelers of model sizes from small to large. Models trained with heterogeneous FAN-TL can achieve at least comparable or even superior performance to ones trained with isomorphic token-labelers. Our large model trained with FAN-TL-B-Hybrid further boosts the clean and robust accuracy to 84.8\% and 69.2\% with the mCE of 42.1\%. This indicates STL is robust to different token-labelers. Such robustness enables us to train a larger student model with a smaller token-labeler, which can reduce the training cost.

\begin{table}[tbp]
\centering
\setlength{\tabcolsep}{3pt}
\resizebox{\linewidth}{!}{
\begin{tabular}{l|cccc}
Model        & Token-Labeler    & IN-1K & IN-C  & mCE ($\downarrow$) \\ \midrule
FAN-S-Hybrid &  - & 83.5      &  64.7    &  47.8   \\
FAN-S-Hybrid & FAN-TL-S-Hybrid &   83.4    &  65.5    &   47.3  \\
FAN-S-Hybrid & FAN-TL-B-Hybrid &   83.3    &  \textbf{65.8}    &   \textbf{46.8}  \\
FAN-S-Hybrid & FAN-TL-L-Hybrid &   \textbf{83.5}    &  65.5   &  47.4   \\ \midrule
FAN-B-Hybrid & - &   83.9   &  66.4    &  45.2   \\
FAN-B-Hybrid & FAN-TL-S-Hybrid &   84.4    &    \textbf{68.5}  &   \textbf{43.2}  \\
FAN-B-Hybrid & FAN-TL-B-Hybrid &   \textbf{84.5}   &  68.2    &  43.6   \\
FAN-B-Hybrid & FAN-TL-L-Hybrid &   84.3    &   68.1   &  43.5   \\ \midrule
FAN-L-Hybrid & - &   84.3    &   68.3   &  43.0   \\
FAN-L-Hybrid & FAN-TL-S-Hybrid &   84.7    &   69.0   &   42.4  \\
FAN-L-Hybrid & FAN-TL-B-Hybrid &   \textbf{84.8}    &   \textbf{69.2}   &  \textbf{42.1}   \\
FAN-L-Hybrid & FAN-TL-L-Hybrid &   84.7    &  68.8    &  42.5   \\ \bottomrule
\end{tabular}
}
\caption{\textbf{Performance comparison of training with different token labelers.} The robustness of student models can be further improved by training with a heterogeneous token labeler.}
\label{tab:heterogeneous_token_labeler}
\vspace{-0.5em}
\end{table}

\begin{table}[htbp]
\centering

\setlength{\tabcolsep}{3pt}
\resizebox{\linewidth}{!}{
\begin{tabular}{l|cccc}
Model       & Token-Labeler    & IN-1K & IN-C & mCE ($\downarrow$) \\ \midrule
FAN-S-Hybrid &  - & \textbf{83.5}      &  64.7    &  47.8   \\
FAN-S-Hybrid & NFNet-F6 (CNN) &   83.2    &  65.8    &   46.9  \\
FAN-S-Hybrid & FAN-TL-B-Hybrid &   {83.3}    &  \textbf{65.8}    &   \textbf{46.8}  \\ \midrule
FAN-B-Hybrid & - &   83.9   &  66.4    &  45.2   \\
FAN-B-Hybrid & NFNet-F6 (CNN) &   83.5   &  67.4    &  44.9   \\
FAN-B-Hybrid &  FAN-TL-S-Hybrid &   \textbf{84.4}    &    \textbf{68.5}  &   \textbf{43.2}  \\ \midrule
FAN-L-Hybrid & - &   84.3    &   68.3   &  43.0   \\
FAN-L-Hybrid & NFNet-F6 (CNN)&   83.9   &  68.3    &  43.2   \\
FAN-L-Hybrid &  FAN-TL-B-Hybrid &   \textbf{84.8}    &   \textbf{69.2}   &  \textbf{42.1}   \\ \bottomrule
\end{tabular}
}
\caption{\textbf{Comparison to the prior SOTA token labeling method.} Our proposed method that employs self-emerging token labels always yields better robustness than NFNet-F6, which validates the benefit of encoding local information of image patches via self-attention.}
\label{tab:comparison_with_NFNet}
\vspace{-0.5em}
\end{table}
\subsection{Comparison to Prior Token Labeling Method} 
\label{exp: comparison_to_prior_tl}
As shown in Sec.~\ref{sec:image_classification_results} and Sec.~\ref{sec:ood_results}, models trained with FAN-TL demonstrate superior performance and robustness than LV-ViTs~\cite{jiang2021all} trained with a CNN token-labeler (i.e., NFNet-F6~\cite{brock2021high}), which may attribute to the self-emerging token labeling and the channel attention design of FAN models. To better understand the effect of token labeling, we train the same FAN models with different methods. Results are summarized in Table~\ref{tab:comparison_with_NFNet}. It can be seen that STL still significantly outperforms the CNN token-labeler even with the same student model. Note that NFNet-F6 has more than 400M parameters and achieves an 86.3\% Top-1 accuracy on ImageNet-1K while the largest FAN-TL (i.e., FAN-TL-L-Hybrid) only has 77.3M parameters and 84.3\% Top-1 accuracy. However, FAN-TL consistently yields better results than NFNet-F6. The rationale behind this is possibly because self-emerging token labels provide self-consistent information to student models.

\subsection{Impacts of Loss Weight}
{Choosing good loss weights is important when multiple losses are jointly optimized. To study the impact of loss weights and verify STL's robustness against different loss weights when training student models, we vary $\beta$ with various values and present the results in Table~\ref{tab:ablation_loss_weights}. We find STL not sensitive to $\beta$. Both clean and robust accuracy fluctuates in a small range. We thus set $\beta=1$ to balance the loss with equal weights for simplicity. Similarly, for the training of token-labelers, $\alpha$ is also set to 1.}

\begin{table}[htbp]
\small
\centering
\setlength{\tabcolsep}{8pt}
\resizebox{\linewidth}{!}{
\begin{tabular}{c|ccccc}
$~~~\beta~~~$ & IN-1K & IN-C & mCE ($\downarrow$) & IN-A & IN-R \\ \midrule
0.5 & 83.5 & 65.5 & 47.3 & 38.5 & 51.7 \\ 
1.0 & 83.4 & 65.5 & 47.3 & 38.2 & 51.8 \\ 
2.0 & 83.5 & 65.6 & 47.2 & 37.3 & 51.1 \\  \bottomrule
\end{tabular}}
\caption{\textbf{Ablation study of loss weight $\beta$.}}
\label{tab:ablation_loss_weights}
\vspace{-0.5em}
\end{table}

\section{Conclusion}
\label{sec:con}

In this paper, we propose a self-emerging token labeling (STL) framework built on Transformer-based models instead of CNNs. STL enables FAN token-labelers to self-produce accurate and semantically meaningful token labels for training student models with dense supervision. Through extensive experiments and ablation studies, we demonstrate that models trained with STL significantly surpass the original FAN counterparts trained only with image-level labels and achieve remarkable robustness improvement in various visual recognition tasks. Our study validates that the self-produced knowledge from ViTs can indeed benefit their pre-training. We hope this work sheds light on the potential and understanding of self-emerging token labeling from ViTs and motivates future research.

{\small
\bibliographystyle{unsrt}
\bibliography{reference}
}

\clearpage
\newpage
\appendix

\section{Appendix}
\subsection{Additional Implementation Details}
We show the detailed training settings in Table~\ref{tab:full_training_settings}. We use multi-node training for FAN base and large models and single-node for tiny and small models. The default batch size and learning rate are adjusted to 1024 and 2e-3 for the single-node training. The training settings are the same for FAN-TL models in the first stage and FAN student models in the second stage.

\begin{table}[htbp]
\centering

\setlength{\tabcolsep}{3pt}
\resizebox{\linewidth}{!}{
\begin{tabular}{l|cccc}
             & STL (T) & STL (S) & STL (B) & STL (L) \\ \hline
Epoch        & 350                & 350               & 350                & 350                \\
Batch Size   & 1024               & 1024              & 2048               & 2048               \\
Weight decay & 0.05               & 0.05              & 0.05               & 0.05               \\
LR           & 2e-3               & 2e-3              & 4e-3               & 4e-3               \\
LR decay     & cosine (0.1)       & cosine (0.1)      & cosine (0.1)       & cosine (0.1)       \\
 \midrule
Dropout      & 0                  & 0                 & 0                  & 0                  \\
Drop path    & 0.05               & 0.25              & 0.35               & 0.45               \\
CutOut prob. & 0.3                & 0.3               & 0.3                & 0.3                \\
CutMix alpha & 1.0                & 1.0               & 1.0                & 1.0                \\
MixUp alpha  & 0.8                & 0.8               & 0.8                & 0.8                \\
RandAug      & 9/0.5              & 9/0.5             & 9/0.5              & 9/0.5              \\
Smoothing    & 0.1                & 0.1               & 0.1                & 0.1                \\ \bottomrule
\end{tabular}}
\caption{Detailed hyper-parameters for training with STL. ``T'', ``S'', ``B'' and ``L'' stand for FAN ``Tiny'', ``Small'', ``Base'' and ``Large'' hybrid models. Note that we use single-node training for the tiny and small models, and thus the total batch size and learning rate are decreased accordingly.}
\label{tab:full_training_settings}
\end{table}

We also try two additional ways to train FAN-TL, but none works well in generating valid token labels. First, we train FAN-TL by only optimizing the loss on the class token (following the conventional training paradigm). Token labels generated by FAN-TL trained with such a method are meaningless and thus can not improve the pre-training. Second, we train FAN-TL by jointly optimizing the losses on the class token and the global average-pooled token while stopping the gradients back-propagating from the global average-pooled token. We aim to evaluate the importance of the gradients that come from the patch tokens side. FAN-TL trained with the second method can not self-identify the misclassified labels for foreground tokens, as the confidence scores of all tokens are high. Therefore, token labels generated by FAN-TL trained with the second method are less accurate. The experiment results indicate that gradients from the class token and global average-pooled token sides are crucial in training FAN-TL.



\subsection{Detailed Results on ImageNet-C}
We comprehensively evaluate model robustness against different types of corruption and summarize the per-category results in Table~\ref{tab:robustness_imc}. For each category, we average the robust accuracy of all five severity levels. The original FAN models already show stronger robustness than other models. Despite this, models trained with STL achieve comparable robustness against blur and noise corruptions and perform exceptionally well against digital and weather corruptions, yielding even higher overall robustness and making them particularly suitable for real-world applications such as autonomous vehicles.

\begin{table*}[htbp]


\centering
\small
\setlength{\tabcolsep}{3pt} 
\resizebox{\linewidth}{!}{
\begin{tabular}{l|c|c|cccc|cccc|cccc|cccc}
\multirow{2}{*}{Model} & \multirow{2}{*}{Param. } & \multirow{2}{*}{Average } & \multicolumn{4}{c|}{Blur} & \multicolumn{4}{c|}{Noise} & \multicolumn{4}{c|}{Digital} & \multicolumn{4}{c}{Weather} \\
\cline{4-19}
& &  & Motion & Defoc & Glass & \multicolumn{1}{c|}{Gauss} & Gauss & Impul & Shot & \multicolumn{1}{c|}{Speck} & Contr & Satur & JPEG & \multicolumn{1}{l|}{Pixel} & \multicolumn{1}{l}{Bright.} & \multicolumn{1}{l}{Snow} & \multicolumn{1}{l}{Fog} & \multicolumn{1}{l}{Frost} \\
\toprule
\multicolumn{18}{c}{Mobile Setting ($<$ 10M)}  \\
\toprule
\multicolumn{1}{l|}{ResNet-18 (\cite{he2016deep})} & 11M & \multicolumn{1}{c|}{32.7 } & 29.6 & 28.0 & 22.9 & \multicolumn{1}{c|}{32.0} & 22.7 & 17.6 & 20.8 & \multicolumn{1}{c|}{27.7} & 30.8 & 52.7 & 46.3 & \multicolumn{1}{c|}{42.3} & 58.8 & 24.1 & 41.7 & 28.2 \\
\multicolumn{1}{l|}{MobileNetV2 (\cite{sandler2018mobilenetv2})} & 4M &  \multicolumn{1}{c|}{35.0} & 33.4 & 29.6 & 21.3 & \multicolumn{1}{c|}{32.9} & 24.4 & 21.5 & 23.7 & \multicolumn{1}{c|}{32.9} & 57.6 & 49.6 & 38.0 & \multicolumn{1}{c|}{62.5} & 28.4 & 45.2 & 37.6 & 28.3 \\
\multicolumn{1}{l|}{EfficientNet-B0 (\cite{tan2019efficientnet})} & 5M & \multicolumn{1}{c|}{41.1 } & 36.4 & 26.8 & 26.9 & \multicolumn{1}{c|}{39.3} & 39.8 & 38.1 & 47.1 & \multicolumn{1}{c|}{39.9} & 65.2 & 58.2 & 52.1 & \multicolumn{1}{c|}{69.0} & 37.3 & 55.1 & 44.6 & 37.4 \\
%
\multicolumn{1}{l|}{PVT-V2-B0 (\cite{wang2021pyramid})} & 3M & \multicolumn{1}{c|}{36.2} & 30.8 & 24.9 & 34.0 & \multicolumn{1}{c|}{35.8} & 33.1 & 35.2 & 44.2 & \multicolumn{1}{c|}{50.6} & 59.3 & 50.8 & 36.6 & \multicolumn{1}{c|}{61.9} & 38.6 & 50.7 & 45.9 & 41.8 \\
\multicolumn{1}{l|}{PVT-V2-B1 (\cite{wang2021pyramid})} & 13M &  \multicolumn{1}{c|}{51.7 } & 45.7 & 41.3 & 30.5 & \multicolumn{1}{c|}{43.9} & 48.1 & 46.2 & 46.6 & \multicolumn{1}{c|}{55.0} & 57.6 & 68.6 & 59.9 & \multicolumn{1}{c|}{50.2} & 71.0 & 49.8 & 56.8 & 53.0 \\
%
\multicolumn{1}{l|}{FAN-T-Hybrid}& 8M & \multicolumn{1}{c|}{{57.4 }} & {52.6}  & {46.7}  & {34.3}  & \multicolumn{1}{c|}{{50.3}} & {55.5}  & {55.8}  & {54.5}  & \multicolumn{1}{c|}{{61.4}} & {65.8}  & {\textbf{73.3}}  & {\textbf{63.8}}  & \multicolumn{1}{c|}{{\textbf{47.9}}} & {74.5}  & {55.0}  & {61.4}  & {52.8}  \\

\multicolumn{1}{l|}{STL (FAN-T-Hybrid)}& 8M & \multicolumn{1}{c|}{{\textbf{58.2} }} & \textbf{52.7}  & {\textbf{48.0}}  & {\textbf{34.9}}  & \multicolumn{1}{c|}{{\textbf{51.3}}} & \textbf{56.8}  & \textbf{56.5}  & \textbf{55.4}  & \multicolumn{1}{c|}{\textbf{61.8}} & {\textbf{66.0}}  & {73.1}  & {62.8}  & \multicolumn{1}{c|}{{47.6}} & \textbf{74.7}  & {\textbf{56.9}}  & {\textbf{64.9}}  & {\textbf{57.2}}  \\
\midrule
%
\multicolumn{18}{c}{GPU Setting (20M+)}  \\
\toprule
\multicolumn{1}{l|}{ResNet-50$^*$ (\cite{he2016deep})} & 25M & \multicolumn{1}{c|}{50.6} & 42.1 & 40.1 & 27.2 & \multicolumn{1}{c|}{42.2} & 42.2 & 36.8 & 41.0 & \multicolumn{1}{c|}{50.3} & 51.7 & 69.2 & 59.3 & \multicolumn{1}{c|}{51.2} & 71.6 & 38.5 & 53.9 & 42.3 \\
%
\multicolumn{1}{l|}{ViT-S (\cite{dosovitskiy2020image})} & 22M & \multicolumn{1}{c|}{54.2} & 49.7 & 45.2 & 38.4 & \multicolumn{1}{c|}{48.0} & 50.2 & 47.6 & 49.0 & \multicolumn{1}{c|}{57.5} & 58.4 & 70.1 & 61.6 & \multicolumn{1}{c|}{57.3} & 72.5 & 51.2 & 50.6 & 57.0 \\
\multicolumn{1}{l|}{DeiT-S (\cite{touvron2021training})}& 22M &  \multicolumn{1}{c|}{58.1} 
& 52.6 & 48.9 & {38.1} & \multicolumn{1}{c|}{51.7} & 57.2 & {55.0} & {54.7} & \multicolumn{1}{c|}{{60.8}} & {63.7} & 71.8 & 64.0 & \multicolumn{1}{c|}{58.3} & 73.6 & {55.1} & 61.1 & {60.7} \\

%
\multicolumn{1}{l|}{FAN-S-Hybrid}& 26M & \multicolumn{1}{c|}{{64.7} } & {{60.8}}  & {\textbf{56.0}}  & {\textbf{44.5}}  & \multicolumn{1}{c|}{{\textbf{58.6}}} & \textbf{65.6}  & \textbf{66.2}  & \textbf{64.8}  & \multicolumn{1}{c|}{\textbf{69.7}} & {67.5}  & {77.4}  & {68.7}  & \multicolumn{1}{c|}{{\textbf{61.0}}} & {78.4}  & {63.2}  & {66.1}  & {62.4}  \\

\multicolumn{1}{l|}{STL (FAN-S-Hybrid)}& 27M & \multicolumn{1}{c|}{\textbf{65.8} } & {\textbf{61.7}}  & {\textbf{56.0}}  & {42.6}  & \multicolumn{1}{c|}{{\textbf{58.6}}} & {65.4}  & {65.6}  & {64.5}  & \multicolumn{1}{c|}{{69.4}} & {\textbf{71.9}}  & \textbf{78.0}  & {\textbf{70.0}}  & \multicolumn{1}{c|}{{59.8}} & \textbf{79.1}  & {\textbf{65.6}}  & {\textbf{71.6}}  & \textbf{65.3}  \\
\midrule

\multicolumn{18}{c}{GPU Setting (50M+)}  \\
\toprule
\multicolumn{1}{l|}{ResNet-101 (\cite{he2019bag})} & 45M & \multicolumn{1}{c|}{59.2 } & 57.0 & 51.9 & 35.6 & \multicolumn{1}{c|}{55.0} & 51.9 & 51.2 & 51.2 & \multicolumn{1}{c|}{61.2} & 67.8 & 75.5 & 67.3 & \multicolumn{1}{c|}{59.9} & 53.6 & 66.2 & 66.4 & 56.4 \\
%

%
\multicolumn{1}{l|}{Swin-S (\cite{liu2021swin})} & 50M & \multicolumn{1}{c|}{60.4} & 56.7 & 51.4 & 34.8 & \multicolumn{1}{c|}{53.4} & 60.1 & 58.4 & 57.8 & \multicolumn{1}{c|}{62.3} & 65.9 & 73.8 & 66.4 & \multicolumn{1}{c|}{62.4} & 76.0 & 55.9 & 67.4 & 60.7 \\
%

%
\multicolumn{1}{l|}{FAN-B-Hybrid}& 50M &  \multicolumn{1}{c|}{{66.4} } & {62.5}  & {58.0}  & {{47.2}}  & \multicolumn{1}{c|}{{60.9}} & {67.6}  & {67.9}  & {67.1}  & \multicolumn{1}{c|}{{71.2}} & {70.8}  & {78.0}  & {69.3}  & \multicolumn{1}{c|}{{62.1}} & {78.9}  & {64.8}  & {69.8}  & {63.3}  \\
\multicolumn{1}{l|}{STL (FAN-B-Hybrid)}& 51M &  \multicolumn{1}{c|}{{\textbf{68.5}} } & {\textbf{65.0}}  & {\textbf{58.7}}  & {\textbf{47.4}}  & \multicolumn{1}{c|}{{\textbf{61.0}}} & {\textbf{69.1}}  & {\textbf{69.3}}  & {\textbf{68.6}}  & \multicolumn{1}{c|}{{\textbf{72.6}}} & {\textbf{73.7}}  & {\textbf{79.1}}  & {\textbf{71.9}}  & \multicolumn{1}{c|}{{\textbf{65.5}}} & {\textbf{80.1}}  & {\textbf{66.8}}  & {\textbf{72.9}}  & {\textbf{67.3}}  \\

\midrule
\multicolumn{18}{c}{GPU Setting (80M+)}  \\
\toprule
\multicolumn{1}{l|}{ViT-B$^*$ (\cite{dosovitskiy2020image})} & 88M &  \multicolumn{1}{c|}{59.7}    & 60.2 & 55.6 & 50.0 & \multicolumn{1}{c|}{57.6} & 54.9 & 52.9 & 53.2 & 62.0 & \multicolumn{1}{c}{52.3} & 71.5 & 68.7 & 71.7 & \multicolumn{1}{c}{74.9} & 52.8 & 57.1 & 41.7  \\
\multicolumn{1}{l|}{DeiT-B (\cite{touvron2021training})} & 89M & \multicolumn{1}{c|}{62.7} & 56.7 & 52.2 & 43.6 & \multicolumn{1}{c|}{55.1} & 64.9 & 63.5 & 61.2 & \multicolumn{1}{c|}{65.7} & 68.2 & 74.6 & 66.9 & \multicolumn{1}{c|}{61.7} & 76.2 & 59.7 & 68.2 & 64.9 \\
\multicolumn{1}{l|}{Swin-B-IN22k (\cite{liu2021swin})} & 88M & \multicolumn{1}{c|}{68.6 } & 66.1 & 62.1 & 48.2 & \multicolumn{1}{c|}{63.2} & 67.3 & 66.2 & 66.4 & \multicolumn{1}{c|}{70.5} & 71.7 & 77.8 & 73.5 & \multicolumn{1}{c|}{74.0} & 80.3 & {66.2} & 74.0 & {66.9} \\
\multicolumn{1}{l|}{ConvNeXt-B (\cite{liu2022convnet})} & 89M &  \multicolumn{1}{c|}{63.6 } & 59.6 & 52.9 & 39.2 & 55.2 & \multicolumn{1}{c}{{65.5}} & 64.8 & 63.7 & 66.7 & \multicolumn{1}{c}{69.9} & 76.2 & 68.9 & {64.6} & \multicolumn{1}{c}{77.8} & 59.2 & {66.7} & 64.3  \\

%
%

%
\multicolumn{1}{l|}{FAN-L-Hybrid} & 77M & \multicolumn{1}{c|}{ {68.3} } &  {65.1} &  {{59.2}} &  {\textbf{49.2}} & \multicolumn{1}{c|}{ {\textbf{61.9}}} & {\textbf{70.1}} &  {\textbf{71.1}} &  {\textbf{69.4}} & \multicolumn{1}{c|}{ {{72.7}}} &  {72.4} &  {77.6} &  {71.8} & \multicolumn{1}{c|}{{\textbf{66.6}}} &  {79.6} &  {65.6} &  {71.3} &  {65.7} \\
\multicolumn{1}{l|}{STL (FAN-L-Hybrid)} & 77M & \multicolumn{1}{c|}{ {\textbf{69.2}} } &  {\textbf{67.1}} &  {\textbf{59.4}} &  {48.6} & \multicolumn{1}{c|}{ {61.7}} & {69.5} &  {71.0} &  {69.1} & \multicolumn{1}{c|}{ {\textbf{73.3}}} &  {\textbf{74.5}} &  {\textbf{79.7}} &  {\textbf{73.2}} & \multicolumn{1}{c|}{{65.6}} &  {\textbf{80.6}} &  {\textbf{66.9}} &  {\textbf{71.6}} &  {\textbf{68.4}} \\
%

\bottomrule
\end{tabular}}
\caption{\textbf{Comparison of model robust accuracy on ImageNet-C (\%)}. Models trained with STL reveal stronger robustness than other models (i.e., models other than FAN) under all corruption categories. They also outperform the original FAN counterparts in most cases (especially against digital and weather corruption) and yield a higher robust accuracy overall. `ResNet-50$^*$' results are reproduced with the same training and augmentation recipes for a fair comparison.}
\label{tab:robustness_imc}
\end{table*}
\subsection{Detailed Results on Cityscapes-C}
Similar to the image classification task, we present the per-category robustness of models on the semantic segmentation task in Table~\ref{tab:benchmark_cityscapes}. We follow the practice in SegFormer~\cite{xie2021segformer} and compute the average mIoU of the first three severity levels for the noise category. For the remaining categories, we compute the average of all five severity levels. Experiment results show that models trained with STL show superior performance in almost all categories than other CNN-based and Transformer-based models, including the original FAN counterparts.

\begin{table*}[htbp]
\centering
\small
\setlength{\tabcolsep}{3pt} 
\resizebox{\linewidth}{!}{
\begin{tabular}{l|c|cccc|cccc|cccc|cccc}
\multirow{2}{*}{Model} & \multirow{2}{*}{Average} & \multicolumn{4}{c|}{Blur} & \multicolumn{4}{c|}{Noise} & \multicolumn{4}{c|}{Digital} & \multicolumn{4}{c}{Weather} \\
\cline{3-18}
&  & Motion & Defoc & Glass & \multicolumn{1}{c|}{Gauss} & Gauss & Impul & Shot & \multicolumn{1}{c|}{Speck} & Bright & Contr & Satur & \multicolumn{1}{l|}{JPEG} & \multicolumn{1}{l}{Snow} & \multicolumn{1}{l}{Spatt} & \multicolumn{1}{l}{Fog} & \multicolumn{1}{l}{Frost} \\
\toprule
%
\multicolumn{1}{l|}{DLv3+ (R50)} & \multicolumn{1}{c|}{36.8} & 58.5 & 56.6 & 47.2 & \multicolumn{1}{c|}{57.7} & 6.5 & 7.2 & 10.0 & \multicolumn{1}{c|}{31.1} & 58.2 & 54.7 & 41.3 & \multicolumn{1}{c|}{27.4} & 12.0 & 42.0 & 55.9 & 22.8 \\
\multicolumn{1}{l|}{DLv3+ (R101)} & \multicolumn{1}{c|}{39.4} & 59.1 & 56.3 & 47.7 & \multicolumn{1}{c|}{57.3} & 13.2 & 13.9 & 16.3 & \multicolumn{1}{c|}{36.9} & 59.2 & 54.5 & 41.5 & \multicolumn{1}{c|}{37.4} & 11.9 & 47.8 & 55.1 & 22.7 \\
%
\multicolumn{1}{l|}{DLv3+ (X65)} & \multicolumn{1}{c|}{42.7} & 63.9 & 59.1 & 52.8 & \multicolumn{1}{c|}{59.2} & 15.0 & 10.6 & 19.8 & \multicolumn{1}{c|}{42.4} & 65.9 & 59.1 & 46.1 & \multicolumn{1}{c|}{31.4} & 19.3 & 50.7 & 63.6 & 23.8 \\
\multicolumn{1}{l|}{DLv3+ (X71)} & \multicolumn{1}{c|}{42.5} & 64.1 & 60.9 & 52.0 & \multicolumn{1}{c|}{60.4} & 14.9 & 10.8 & 19.4 & \multicolumn{1}{c|}{41.2} & 68.0 & 58.7 & 47.1 & \multicolumn{1}{c|}{40.2} & 18.8 & 50.4 & 64.1 & 20.2 \\
\midrule
\multicolumn{1}{l|}{ICNet (\cite{zhao2018icnet})} & \multicolumn{1}{c|}{28.0} & 45.8 & 44.6 & 47.4 & \multicolumn{1}{c|}{44.7} & 8.4 & 8.4 & 10.6 & \multicolumn{1}{c|}{27.9} & 41.0 & 33.1 & 27.5 & \multicolumn{1}{c|}{34.0} & 6.3 & 30.5 & 27.3 & 11.0 \\
\multicolumn{1}{l|}{FCN8s (\cite{long2015fully})} & \multicolumn{1}{c|}{27.4} & 42.7 & 31.1 & 37.0 & \multicolumn{1}{c|}{34.1} & 6.7 & 5.7 & 7.8 & \multicolumn{1}{c|}{24.9} & 53.3 & 39.0 & 36.0 & \multicolumn{1}{c|}{21.2} & 11.3 & 31.6 & 37.6 & 19.7 \\
\multicolumn{1}{l|}{DilatedNet (\cite{yu2015multi})} & \multicolumn{1}{c|}{30.3} & 44.4 & 36.3 & 32.5 & \multicolumn{1}{c|}{38.4} & 15.6 & 14.0 & 18.4 & \multicolumn{1}{c|}{32.7} & 52.7 & 32.6 & 38.1 & \multicolumn{1}{c|}{29.1} & 12.5 & 32.3 & 34.7 & 19.2 \\
\multicolumn{1}{l|}{ResNet-38} & \multicolumn{1}{c|}{32.6} & 54.6 & 45.1 & 43.3 & \multicolumn{1}{c|}{47.2} & 13.7 & 16.0 & 18.2 & \multicolumn{1}{c|}{38.3} & 60.0 & 50.6 & 46.9 & \multicolumn{1}{c|}{14.7} & 13.5 & 45.9 & 52.9 & 22.2 \\
\multicolumn{1}{l|}{PSPNet (\cite{zhao2017pyramid})} & \multicolumn{1}{c|}{34.5} & 59.8 & 53.2 & 44.4 & \multicolumn{1}{c|}{53.9} & 11.0 & 15.4 & 15.4 & \multicolumn{1}{c|}{34.2} & 60.4 & 51.8 & 30.6 & \multicolumn{1}{c|}{21.4} & 8.4 & 42.7 & 34.4 & 16.2 \\
\multicolumn{1}{l|}{ConvNeXt-T (\cite{liu2022convnet})} & \multicolumn{1}{c|}{54.4} & 64.1 & 61.4 & 49.1 & \multicolumn{1}{c|}{62.1} & 34.9 & 31.8 & 38.8 & \multicolumn{1}{c|}{56.7} & 76.7 & 68.1 & 76.0 & 51.1 & \multicolumn{1}{c}{25.0} & 58.7 & 74.2 & 35.1  \\
\midrule
\multicolumn{1}{l|}{SETR (DeiT-S) (\cite{zheng2021rethinking})}& \multicolumn{1}{c|}{{55.5}} & {61.8}  & {61.0}  & {59.2}  & \multicolumn{1}{c|}{{62.1}} & {36.4}  & {33.8}  & {42.2}  & \multicolumn{1}{c|}{{61.2}} & {73.1} &{63.8}  & {69.1}   & \multicolumn{1}{c|}{{49.7}}   & {41.2}  & {60.8} & 63.8  & {32.0}  \\
\multicolumn{1}{l|}{Swin-T (\cite{liu2021swin})}& \multicolumn{1}{c|}{{47.5}} & {62.1}  & {61.0}  & {48.7}  & \multicolumn{1}{c|}{{62.2}} & {22.1}  & {24.8}  & {25.1}  & \multicolumn{1}{c|}{{42.2}} & {75.8} & {62.1}  & {75.7}   & \multicolumn{1}{c|}{{{33.7}}}   & {19.9}  & 56.9 & {72.1}  & {30.0}  \\
\multicolumn{1}{l|}{SegFormer-B0 (\cite{xie2021segformer})}& \multicolumn{1}{c|}{{48.9}} & {59.3}  & {58.9}  & {51.0}  & \multicolumn{1}{c|}{{59.1}} & {25.1}  & 26.6 & {30.4}  & {50.7}  & \multicolumn{1}{c}{{73.3}} & 66.3 & {71.9}  & {31.2}    & \multicolumn{1}{c}{{22.1}} & {52.9}  & {65.3}  & {31.2}  \\
\multicolumn{1}{l|}{SegFormer-B1 (\cite{xie2021segformer})}& \multicolumn{1}{c|}{{52.6}} & {63.8}  & {63.5}  & {52.0}  & \multicolumn{1}{c|}{{29.8}} & {23.3}  & 35.4 & {56.2}  & {76.3}  & \multicolumn{1}{c}{{70.8}} & 74.7 & {36.1}  & {56.2}    & \multicolumn{1}{c}{{28.3}} & {60.5}  & {70.5}  & {36.3}  \\%
\multicolumn{1}{l|}{SegFormer-B2 (\cite{xie2021segformer})}& \multicolumn{1}{c|}{{55.8}} & {68.1}  & {67.6}  & {58.8}  & \multicolumn{1}{c|}{{68.1}} & {23.8}  & 23.1 & {27.2}  & {47.0}  & \multicolumn{1}{c}{{79.9}} & 76.2 & {78.7}  & {46.2}    & \multicolumn{1}{c}{{34.9}} & {64.8}  & {76.0}  & {42.1}  \\
%
\midrule
%

%
\multicolumn{1}{l|}{FAN-B-Hybrid}& \multicolumn{1}{c|}{ {66.9}} & {70.0}  & {69.0}  & {64.3}  & \multicolumn{1}{c|}{{70.3}} &  {55.9}  &  {60.4}  &  {61.1}  & \multicolumn{1}{c|}{ {70.9}} &  {81.2} & {76.1}  &  {80.0}  & {57.0}  & \multicolumn{1}{c}{{54.8}}   & {72.5}  &  {78.4}  & {52.3}  \\

\multicolumn{1}{l|}{STL (FAN-B-Hybrid)}& \multicolumn{1}{c|}{ {68.6}} & {70.1}  & {71.0}  & {\textbf{66.4}}  & \multicolumn{1}{c|}{{\textbf{71.9}}} &  {58.6}  &  {62.3}  &  {63.8}  & \multicolumn{1}{c|}{ {73.0}} &  {81.5} & {77.4}  &  {80.6}  & {\textbf{62.4}}  & \multicolumn{1}{c}{{54.1}}   & {71.7}  &  {79.3}  & {53.2}  \\
\multicolumn{1}{l|}{FAN-L-Hybrid}& \multicolumn{1}{c|}{{68.7}} & {70.0}  & {69.9}  & {65.3}  & \multicolumn{1}{c|}{{71.6}} & \textbf{60.0}  & {64.5}  & {63.3}  & \multicolumn{1}{c|}{{71.6}} & {81.4} & {76.2}  & {80.1}  & {62.3}  & \multicolumn{1}{c}{{53.1}}   & {73.9}  & {78.9}  & \textbf{54.4}  \\

\multicolumn{1}{l|}{STL (FAN-L-Hybrid)}& \multicolumn{1}{c|}{ {\textbf{69.2}}} & {\textbf{71.4}}  & {\textbf{70.0}}  & {66.1}  & \multicolumn{1}{c|}{{70.7}} &  {58.7}  &  {\textbf{66.8}}  &  {\textbf{65.1}}  & \multicolumn{1}{c|}{ {\textbf{74.8}}} &  {\textbf{81.9}} & {\textbf{77.3}}  &  {\textbf{81.3}}  & {58.1}  & \multicolumn{1}{c}{\textbf{55.2}}   & {\textbf{74.5}}  &  {\textbf{79.9}}  & {53.6}  \\
\bottomrule
\end{tabular}
}
\caption{\textbf{Comparison of model robust accuracy on Cityscapes-C (\%)}. Models trained with STL show stronger robustness in almost all corruption categories than other CNN-based and Transformer-based models (including the original FAN models). ``DLv3+'' stands for DeepLabv3+ \cite{chen2018encoder}. The mIoUs of CNN models are replicated from \cite{kamann2020benchmarking}. 
}
\label{tab:benchmark_cityscapes}
\end{table*}

\subsection{Visualization of Token Labels}
\begin{figure*}[htbp]
    \centering
    \resizebox{1\textwidth}{!}{
    \includegraphics{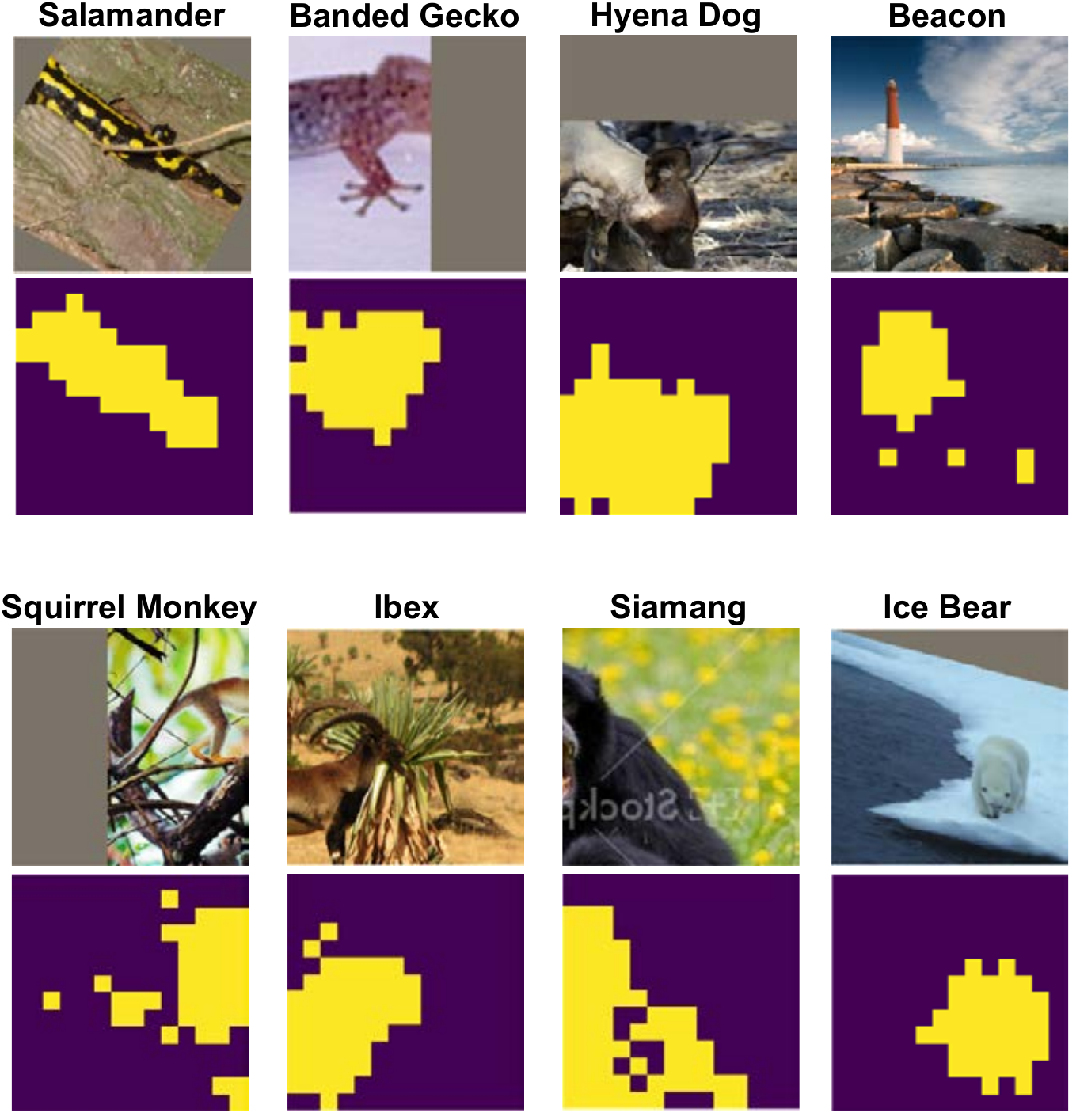}}
    \caption{\textbf{More visualization results of token labels generated by FAN-TL.} FAN-TL performs consistently well in capturing the object gestalt and generating accurate token labels for images with spatial-only data augmentations.}
    \label{fig:tl_visualization}
\end{figure*}

As shown in Fig.~\ref{fig:token_confidence}, FAN-TL can generate semantically meaningful token labels. By applying the spatial-only data augmentation and Gumbel-Softmax, we further retain  more accurate token labels of the target object. We demonstrate more visualization results of token labels generated by FAN-TL with spatial-only data augmentation in Fig~\ref{fig:tl_visualization}. It can be seen that FAN-TL performs consistently well in capturing the object gestalt and generating accurate token labels for images with rotation, crop, shear and translation.

\end{document}